# Lexical-semantic resources: yet powerful resources for automatic personality classification

Xuan-Son Vu [†1*], Lucie Flekova [¶2], Lili Jiang [†3], Iryna Gurevych [§4]

¶ Amazon Research Germany, Aachen
†Department of Computing Science, Umeå University, Sweden
§UKP Lab, Computer Science Department, Technische Universität Darmstadt
[†1]sonvx@cs.umu.se, [¶2]Lflekova@amazon.com
[†3]lili.jiang@cs.umu.se, [§4]gurevych@ukp.informatik.tu-darmstadt.de

## Abstract

In this paper, we aim to reveal the impact of lexical-semantic resources, used in particular for word sense disambiguation and sense-level semantic categorization, on automatic personality classification task. While stylistic features (e.g., part-of-speech counts) have been shown their power in this task, the impact of semantics beyond targeted word lists is relatively unexplored. We propose and extract three types of lexical-semantic features, which capture high-level concepts and emotions, overcoming the lexical gap of word n-grams. Our experimental results are comparable to state-of-the-art methods, while no personality-specific resources are required.

## 1 Introduction

Automatic personality classification (APC) has been employed on *user generated content* (UGC), such as Tweets, to collect the user personality for various personalized intelligent applications, including recommender systems (Hu and Pu, 2011), mental health diagnosis (Uba, 2003), recruitment and career counseling (Gardner et al., 2012). Especially, the recommender applications benefit from knowing the personality of real as well as fictional characters (Flekova and Gurevych, 2015). For example, if a user is known to favor the personality traits displayed by the main characters of, say, *Terminator* [1] and *Rambo*[1], then the system should automatically recommend movies with similar characters.

Currently, the performance of APC depends on how user personality is modeled and what types of personality features can be extracted. Regarding the first factor, one well-known model called Five Factor Model (Costa and McCrae, 2008) has been highly accepted as a standard model. It consists of five personality traits (i.e., extraversion, neuroticism, agreeableness, conscientiousness, openness to experience). The APC task is then formulated as a regular document classification on these five labels. To the second factor of feature extraction, the existing studies heavily depend on personality specific resources such as linguistic inquiry word count (LIWC) (Pennebaker et al., 2007). These resources, however, are rather time consuming and expensive to construct especially for minor languages (Vu and Park, 2014). Moreover, the resource construction requires expertise in both psychology and linguistic (e.g., LIWC). In contrast, it is observed that lexical-semantic features which could be extracted from the publicly available lexical resources (e.g., WordNet (Miller, 1995)) can help to improve the performance of the APC task. However, their impact on real world UGC data for APC had been relatively unexplored.

Among lexical-semantic features, sense-level features were explored in previous works (Kehagias et al., 2003; Vossen et al., 2006) with varying conclusions. In this paper, we conduct extensive experiments, aiming at obtaining a more detailed understanding of whether or not the senses can be beneficial in certain cases compared to word-based fea-tures. Broadly, we explore the use of word senses, supersenses, and WordNet sentiment features (Baccianella et al., 2010) in personality classification. Our main contributions are:

- Investigating the impact of different lexical-semantic features on APC task.

- Revealing the accumulated benefit by combining word sense disambiguation (WSD) with semantic and sentiment features in APC.

- Proposing and evaluating a feature selection method called *Selective.WSD* to improve WSD usage in APC.

- Proposing a unified framework on top of the UIMA framework [2] to integrate different lexical-semantic resources for APC.

---

*The research by the 1st and the 2nd authors has been done during their employment at the UKP Lab, Technische Universität Darmstadt, Germany, and supported by the German Research Foundation under grant No. GU 798/14-1.

[1] Famous fiction/action movies.

The rest of this paper is organized as follows. Section 2 presents the related work and our novel contributions, as well as background knowledge of the Five Factor Model. Section 3 describes the experimental datasets. Our proposed framework and methodology are presented in Section 4. Experimental results and discussion are in Section 5. Section 6 concludes this paper.

## 2 Related Work and Background

Previous studies concerned the positive impact of sense-level features (i.e., using Word-Net based WSD) on the performance of document classification systems (Rose et al., 2002; Kehagias et al., 2003; Moschitti and Basili, 2004; Vossen et al., 2006). Though they had different focuses, they suggest that word senses are not adequate to improve text classification accuracy. Vossen et al. (2006) report an improvement from 0.70 to 0.76 F-score while negative results have been reported by Kehagias et al. (2003). This is why supersenses, the coarse-grained semantic labels based on WordNet's lexicographer files, have recently gained attention for text classification tasks. In this paper, we further explore the impact of these features in personality prediction.

There have been many different attempts to automatically classify personality traits from texts. However, there were not any studies incorporating senses, supersenses, and sentiment features into the APC. Some works (Iacobelli et al., 2011; Bachrach et al., 2012; Iacobelli and Culotta, 2013; Okada et al., 2015) start from the data and seek linguistic cues associated with personality traits, while other approaches (Mairesse et al., 2007; Golbeck et al., 2011; Farnadi et al., 2016) make heavy use of external resources, such as LIWC (Pennebaker et al., 2007), MRC (Wilson, 1988), NRC (Mohammad et al., 2014), *SentiStrength* [3], where they detect the correlations between those resources and personality traits.

However, the resources require the efforts of experts in psychology and linguistics, e.g., LIWC of Pennebaker et al. (2007), to construct. This constrains the available resources for APC, especially for minor languages. Thus, we aim at broadly available resources (e.g., WordNet and SentiWordNet), to benefit APC.

Close to our work, Mairesse et al. (2007) run personality prediction in both observer judgments through conversation and self-assessments using text via the Five Factor Model. They also exploit two lexical resources as features, LIWC and MRC, to predict both personality scores and classes using Support Vector Machines (SVMs) and M5 trees respectively. As for personality prediction on social network data, Golbeck et al. (2011) use both linguistic features (from LIWC) and social features (i.e., friend count, relationship status). Recently, Farnadi et al. (2016) deal with the automatic personality classification based on users social media traces, which include three of the four datasets in our study. However, similar to other studies (Mairesse et al., 2007; Farnadi et al., 2013), they mainly use the personality specific resources.

At the time of writing, the use of personality specific resources for APC has received much attention, while the impact of lexical-semantic features has been neglected. The only existing work that explores sense-level features is from Flekova and Gurevych (2015). They partially used sense-level features among others (i.e., lexical features, stylistic features, and word embedding features) for personality profiling of fictional characters. As a complement of the existing work on automatic personality classification, the novel contributions of this paper include: (1) we present how WSD and lexical-semantic features influence personality prediction by conducting different experiments on four public datasets; and (2) we explore the accumulated impact of supersenses and sentiment features in combination with WSD.

---

[2]https://uima.apache.org/

[3]http://sentistrength.wlv.ac.uk/

**The Five Factor Model**

In personality prediction, the most influential Five Factor Model (*FFM*) has become a standard model in psychology over the last 50 years (Mairesse et al., 2007). The five factors are defined as extraversion, neuroticism, agreeableness, conscientiousness, and openness to experience. Pennebaker and King (1999) identify many linguistic features associated with each of personality traits in *FFM*. (1) Extroversion (cEXT) tends to seek stimulation in the external world, the company of others, and to express positive emotions. (2) Neurotics (cNEU) people use more 1st person singular pronouns, more negative emotion words than positive emotion words. (3) Agreeable (cAGR) people express more positive and fewer negative emotions. Moreover, they use relatively fewer articles. (4) Conscientious (cCON) people avoid negations, negative emotion words and words reflecting discrepancies (e.g., should and would). (5) Openness to experience (cOPN) people prefer longer words and tentative expressions (e.g., perhaps and maybe), and reduce the usage of 1st person singular pronouns and present tense forms.

Table 1: A quick overview of the four datasets with the number of sentences (#Sen), the number of words (#Word), and the number of users (#Users). *Non-standard words* may be either out-of-vocabulary tokens (e.g., *tmrw* for 'tomorrow') or in-vocabulary tokens (e.g., *wit* for *with* in 'I come wit you').

| Dataset | #Sen | #Word | #Users | Non-standard words |
|---|---|---|---|---|
| TWITTER | 145.7 | 216.8 | 153 | 51.27% |
| FACEBOOK | 67.1 | 78.3 | 250 | 23.3% |
| ESSAYS | 48.8 | 15.3 | 2469 | 30.85% |
| YOUTUBE | 41.7 | 29.5 | 404 | 8.05% |

## 3 Dataset and Statistics

### 3.1 Dataset Overview

We conducted our experimental studies on four public datasets, three of which are from public social media platforms (i.e., Twitter, Facebook, Youtube) and the fourth one is a well-known public dataset specially for personality research. These datasets are chosen for their popularity and diversity in data size, scale of users, and writing styles.

- TWITTER : collected by PAN' 15 (Stamatatos et al., 2015), it contains Tweets of 328 Twitter users in 4 languages in which only the Tweets come from 153 users written in English are selected in this study.

- FACEBOOK : collected through the myPersonality project [4] (Stillwell and Kosinski, 2015) containing status updates of 250 Facebook users with 9,917 status updates and personality labels.

- YOUTUBE : collected by Biel et al. (2011), it consists of a collection of behavioral features, speech transcriptions, and personality impression scores for a set of 404 YouTube vloggers. About 28 hours of video were annotated.

- ESSAYS : collected and analysed by Pennebaker and King (1999). It contains 2,479 essays from psychology students, who were required to write whatever came into their mind for 20 minutes. The data includes users, raw text, and gold standard classification labels.

### 3.2 Data Statistics

Table 1 shows the overview statistics of the four datasets. All values are normalized by the number of users in each corresponding dataset. *Non-standard words* denotes the fraction of non-standard words (unseen vocabularies in WordNet) over the total number of words in each dataset.

The statistics in Table 1 indicate that Twitter dataset has the highest value of #Sen and #Word but the lowest number of users. Moreover, the TWITTER dataset also has the highest ratio of *non-standard words*, which makes it more challenges to the APC task. All in all, these diverse characteristics benefit our results analysis on improving personality classification.

---
[4]http://myPersonality.com

As depicted in Figure 1, we design a system based on UIMA framework[5] for experimental studies. It contains three main processes including (1) Data Loading and Data Processing, 2) Feature Extraction, (3) Personality Classification and Evaluation. After loading data into the whole system (i.e., four datasets and lexical resources), feature extraction is performed. Afterwards, we formulate personality classification as a binary classification on each personality trait since more than one trait can be embodied in a user. We apply the SVM classifier (linear kernel) and the TF-IDF feature weighting scheme. In the evaluation, we use 10-fold cross validation, i.e., rotating the 10% test data selection over the dataset and training the SVM classifier on the 90% of not-tested data, to get accuracy scores. Since the goal of this paper is revealing the impact of different lexical-semantic features in APC, we used exactly the same classification algorithm as used in the popular work of Mairesse et al. (2007). Details about the second process of feature extraction will be described in the following subsection.

### 3.3 Feature Extraction

Based on our observations and the previous studies, we found that people with different personal traits have different writing styles and word usage. For example, *neurotic* and *extrovert* people use the emotion words significantly differently. *Neurotic* people use more 1st person single pronouns while less positive emotional words. And it is observed that *openness* people use more abstract concepts. Motivated by these observations, we manage to capture these personality trait differences by extracting the semantic and sentiment features.

## 4 Methodology

We denote four kinds of features as F = {WORD, SENSE, S_SENSE, SENTI} where WORD is a set of word-level features, SENSE is a set of sense-level features, S_SENSE is a set of

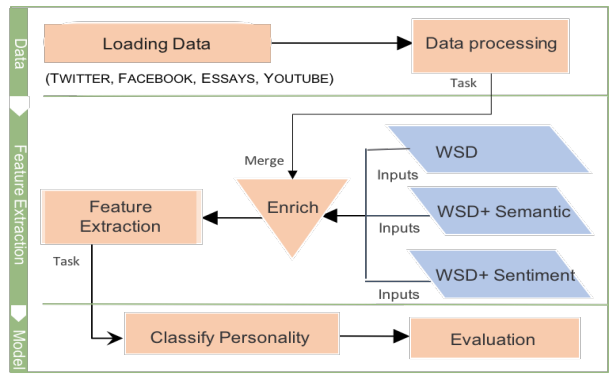

Figure 1: Workflow of the experimental pipeline.

WordNet supersense features, and SENTI is a set of sentiment features. (S_SENSE) is extracted from WordNet supersenses as a complement to SENSE. Regarding sense-level feature, we applied two different WordNet based WSD algorithms, SimLesk and MostFreq (Miller et al., 2013). Correspondingly, instead of SENSE, we have two different feature sets WN-S-LESK and WN-MFS. Thus, we finally have the feature list of F = {WORD, WN-S-LESK, WN-MFS, S_SENSE, SENTI}

**Semantic Features**

Regarding semantic features, we focus on extracting topic information given input texts from different people. We firstly recognize lexical knowledge by applying WordNet semantic labels[6]. For example, based on the given personal texts, after extracting word n-grams, the topic information is detected and organized in the form of *pos.suffix*. Here, *pos* denotes part-of-speech and *suffix* organizes groups of synsets into different categories (e.g., *a tiger* can be categorized into *noun.animal* and *a tree* is categorized into *noun.plant*). In this paper, DKPro Uby (Gurevych et al., 2012) is further employed to extract all above required information to represent in *pos* and *suffix* from given texts.

**Sentiment Features**

For sentiment features, we extracted emotional information, which are extremely important to characterize personality according to Pennebaker and King (1999). For example, neurotics use more negative emotion words (e.g., *ugly* and *hurt*) than positive emotion (e.g., *happy* and *love*). In details, we applied the sentiment word disambiguation algorithm (i.e., SentiWordNet) to match the disambiguated word senses for each term with three scores, Positive (*P*), Negative (*N*) and Objective/Neutral (*O*) scores. Finally, we obtained the individually final *P*, *N* and *O* scores for each personal text, which were averaged by the total number of sentiment features.

---

[5] https://uima.apache.org/

[6] https://wordnet.princeton.edu/man/lexnames.5WN.html

### 4.1 Word Sense Disambiguation

Above, we have discussed and presented feature extraction for APC. However, one primary challenge in feature extraction is word sense ambiguity. To address this challenge, word sense disambiguation (WSD) is broadly applied to match the exact sense of an ambiguous word in a particular context. For word, sense, supersense, and sentiment features, it is necessary to first disambiguate the words to reduce the semantic gap.

However, due to the high ambiguity of words, it is extremely challenging to detect the exact sense in a certain context. Postma et al. (2016) showed that current WSD systems perform an extremely poor performance on low frequent senses. To address this challenge, we propose an algorithm *Selective.WSD* to reduce the side effect of WSD by finding senses of a word subset rather than all possible words in the BoW model. *Selective.WSD* is presented in Algorithm 1. The algorithm takes a word-level document as an input to return a mixture of word-level and sense-level feature list. The *wordLevelFeature(f)* function in the algorithm will return a word-level feature (e.g., bank) of a sense-level feature (e.g., bank%1) by removing the extra notation (e.g., %1). The function of *wsd.annotateSenses* in the algorithm is implemented based on DKPro WSD (Miller et al., 2013) - annotating the exact sense of a disambiguated word in a context. In the following experimental study section, we will show the impact of WSD on personality prediction.

### 4.2 Feature Selection

Feature selection is naturally motivated by the need to automatically select the best determinants for each personality trait. Thus, we can derive a qualitative description of the state characteristics. In this way, the noisy features are filtered out. We used the $\chi^2$ feature selection algorithm before feeding the features (i.e., word, sense, supersense, and sentiment features) to a classifier. The feature selection strategy was chosen empirically based on our preliminary experiments on training dataset, where we compared $\chi^2$ with three other state-of-the art feature selection methods for the supervised classification (i.e., Information Gain, Mutual Information, and Document Frequency thresholding (Yang and Pedersen, 1997)), and $\chi^2$ outperformed.

**Procedure 1** *Selective.WSD*
**Input:** a word-level document.
**Output:** a selective mixture of word-level and sense-level feature list.
1: *featuresL* ← initialize an empty list
2: *L* ← topK word-level features ordered by $\chi^2$
3: **for** sentence *s* ∈ d*ocument d* **do**
4:    *mixFeatList* ← *wsd.annotateSenses*(*s*)
5:    **for** feature *f* ∈ *mixFeatList* **do**
6:      **if** *wordLevelFeature*(*f*) ∉ *L* **then**
7:        *f* ← *wordLevel*(*f*)
8:      **else**
9:        *f* ← *senseLevel*(*f*)
10:    *featuresL* ←⊔ *f*
   **return** *featuresL*

Table 2: Abbreviation list of the feature set

| ID | Description |
|---|---|
| WORD | Word-level features. |
| WN-WORD | Word-level features in which only words that present in WordNet are used. |
| WN-MFS | Sense-level features based on the most frequent sense algorithm. |
| WN-S-LESK | Sense-level features based on the Simplified Lesk algorithm. |
| S_SENSE | WordNet semantic label (or WordNet supersense) features. |
| SENTI | Three sentiment features including posscore, negscore, and neuscore. |

## 5 Experiment and Analysis

We conducted extensive experiments to investigate the impact of different lexical-semantic features on the APC task. All the feature abbreviations we use are listed in Table 2.

## 5.1 Experiment Settings

We compared four pipelines based on different lexical-semantic feature settings. In the first and simplest pipeline, the documents are segmented into words used as features. We further refer to this setup as WORD. The subsequent feature selection and classification, specified below, is the same for all pipelines. In the second processing pipeline, the documents are segmented to words, and the words are further annotated with their part-of-speech and lemma. With these annotations, we can look them up in WordNet. Only those words, which are present in WordNet, are then used as bag-of-words features. This intermediate step reveals which changes in performance can be attributed to the lexicon coverage as opposed to the WSD quality. We refer to this setup as WN-WORD. The third processing pipeline is similar to the previous one, but after the WN-WORD lookup step performed, in addition, the WordNet based WSD is employed to extract sense-level features. For each of the words present in WordNet, the resulting sense and its WordNet semantic label (S_SENSE) are both used as two features. There are two possible configurations in the third pipeline, which differ in the WSD algorithm used (see subsection 4.1). We experimented with the most frequent sense baseline (denoted further as *WN-MFS-S_SENSE*) and Simplified Lesk algorithm (*WN-S-LESK-S_SENSE*). Differently from the third pipeline, in the fourth pipeline, for each sense, we calculate three sentiment scores (positive, negative, neutral) by applying SentiWordNet and add them as three extra features. We refer to this setup as *WN-S-MFS-S_SENSE-SENTI* and *WN-S-LESK-S_SENSE-SENTI* for the Most Frequent Sense and the Simplified Lesk algorithm correspondingly. All results from the above four different pipelines are shown in Figure 2 and Figure 3. More discussions are present in the following subsections.

## 5.2 Experimental Result Demonstration

As shown in Figure 2 and 3, though the APC performance of different configurations varies on different datasets, we have some interest-ing observations. For example, for predicting conscientiousness, openness and agreeableness personality traits, using the WSD algorithm always decreases the performance across all datasets, while the prediction performance on extraversion and neuroticism improves 75% cases. The restriction to WordNet-only words is helpful in 10/24 ≈ 41% of the cases, especially on ESSAYS dataset. It is noteworthy that the *S-LESK* related settings (i.e., *S-LESK-S_SENSE* and *S-LESK-S_SENSE-SENTI*) perform better than *MFS* related settings (i.e., *MFS-S_SENSE* and *MFS-S_SENSE-SENTI*).

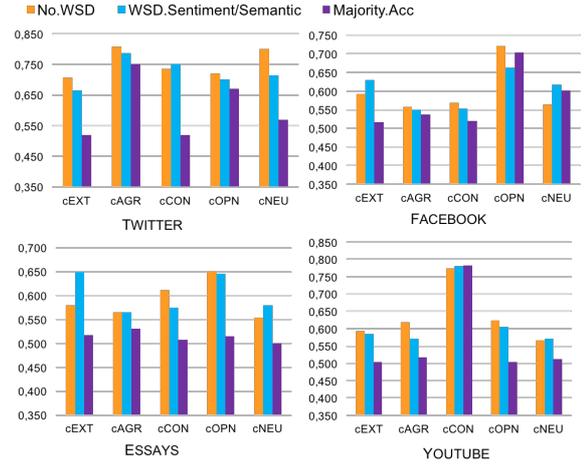

Figure 2: A comparison between not-using WSD (i.e, No.WSD) versus using WSD in a combination with sentiment/semantic features (i.e., WSD.Sentiment/Semantic) in the four datasets. The majority accuracy (i.e., Majority.Acc) is the accuracy when we predict all test instances to a major class.

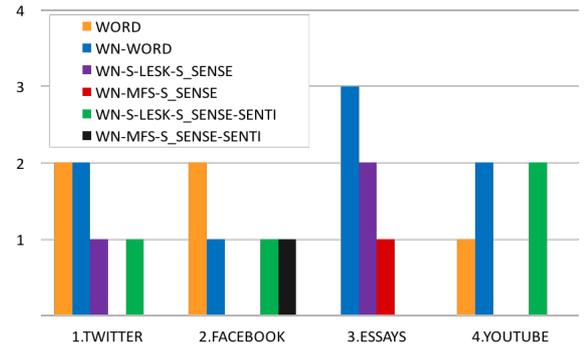

Figure 3: The overall number of times that each feature setting achieves the best performance in the four datasets.

## 5.3 Experimental Result Analysis

For the classification results, we have the following two observations: a) The restriction to WordNet words (WN-WORD vs. WORD) helps the most datasets (3 out of 4 datasets) for predicting openness and agreeableness. b) The positive effects of SENTI features on predicting neuroticism (2 out of 4 datssets). Detailed analysis are presented in the following paragraphs.

**Impact of word feature (WORD)**

We observe that in the all-words approach, there are many pronouns in the top-ranked features. The pronouns are later removed when filtering for WordNet words only. The experimental results show that removing these high-ranked features (e.g., pronouns, particles, and punctuation) increases the accuracy on ESSAYS dataset in all cases, while for other three datasets the feature impact varies based on different data. One possible explanation is that the essays are written in a more thoughtful manner, focused on the inner thoughts. They may, therefore, carry more personality-related information in the content words than the social media data, where the interjection and smileys are more revealing than the topic under discussion. Restriction to WordNet words only thus helps in the essays to better represent the document.

**Impact of sentiment feature (SENTI)**

In the WSD-S_SENSE-SENTI setup, a better result is achieved on cNEU label since *neuroticism* people tend to use more emotional words (Pennebaker and King, 1999).

**Comparison with the state-of-the-art results**

Table 3: Performance in comparison with the state-of-the-art results on the FACEBOOK dataset.

| Trait | Majumder et al. (2017) | Ours (Majority.Acc) |
|---|---|---|
| cOPN | 62.68 | **72.10** (70.40) |
| cCON | 57.30 | 56.80 (52.00) |
| cEXT | 58.09 | **62.10** (38.40) |
| cAGR | 56.71 | 55.80 (53.60) |
| cNEU | 59.38 | **61.70** (39.60) |
| Avg | 58.83 | 58.64 (50.80) |

Given our purpose is not about competing for performance but rather exploring the effectiveness of the general lexical-resources in APC. However, in Table 3, we draw a comparison with the recent best results of Majumder et al. (2017) to show that we get very competitive results on the FACEBOOK dataset. This is a very fair comparison since Majumder at al. used exactly the same evaluation settings as ours. It is worth to mention that, Majumder et al. (2017) used complex neural network models while we used the simple SVM model without tuning parameters. For other datasets, it is difficult to show a fair comparison since previous works (e.g., Farnadi et al. (2016)) regard the APC task as a linear regression problem instead of classification.

## 5.4 Discussion on Different Pipeline Settings

Figure 3 shows the ratio of the number of times each feature setting achieves the best performance over other pipelines in each dataset. In the picture, we can see the WN-WORD setting works well most of the time across four datasets. Therefore, the restriction to WordNet words is a low-cost and effective process to improve personality prediction.

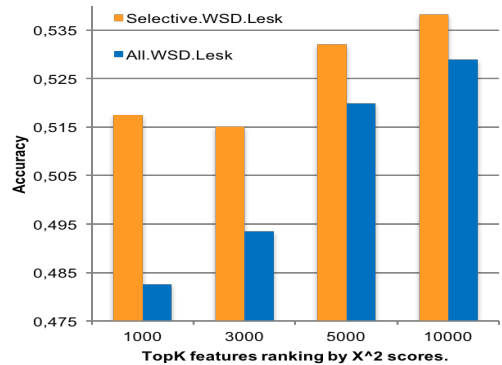

Figure 4: A test on cEXT personal trait of ESSAYS dataset to compare between Selective.WSD and All.WSD.

**Impact of WSD on APC**

We found that the WSD does not generally lead to an improvement in classification results except arbitrary dataset-specific differences, which can be largely attributed to the lemmatization and POS tagging. However, in contrary to previous beliefs (Sanderson, 1994; Gonzalo et al., 1998), the performance of the WSD algorithms is not the major issue for stagnating performance. Rather, it is the reduction of the representative scope of bag-of-words (since function words are not present in the lexicon) and the reduction of the impact of multi-POS words (since those are assigned different senses), which leads to a lower ranking of otherwise highly predictive features. For example, in table 4, in the WN-WORD setup, the word *worry* is ranked to predict *extraversion* with $\chi^2$ = .007, while the sense $worry_1v$ is ranked to predict *introversion*, i.e., the opposite class of *extraversion*, with $\chi^2$ = -.004. Furthermore, as pointed out in (Gale et al., 1992), if a polysemous word appears two or more times in a discourse, it is likely that all the occurrences will share the same coarse-grained sense. A fine-grained WSD might be therefore counter-productive. However, while the effect of WSD itself in a BoW setup is marginal, we observe that the WSD quality is rather high. This implies that the assigned senses can be reliably used to query additional information about the word meaning (and relations to other words) from the lexical-semantic resources.

| WORD | $\chi^2$ | WN-WORD | $\chi^2$ |
|---|---|---|---|
| love | .012 | love | .026 |
| boyfriend | .008 | music | .010 |
| 'd | .008 | sleep | .009 |
| me | .007 | assignment | .009 |
| so | .006 | proud | .008 |
| people | .006 | boyfriend | .007 |
| much | .005 | worry | .007 |
| we | .005 | people | .007 |
| thinks | .005 | awkward | .007 |

| WN-MFS | $\chi^2$ | WN-S-LESK | $\chi^2$ |
|---|---|---|---|
| $love_1v$ | .016 | $love_1v$ | .017 |
| $music_1n$ | .009 | $assignment_1n$ | .009 |
| $guy_1n$ | .009 | $sleep_1v$ | .008 |
| $good_1a$ | .009 | $street_4n$ | .007 |
| $proud_1a$ | .008 | $love_1n$ | .006 |
| $assignment_1n$ | .008 | $sleep_1n$ | .006 |
| $boyfriend_1n$ | .008 | $music_1n$ | .005 |
| $real_1a$ | .006 | $good_6a$ | .005 |
| $sleep_1v$ | .006 | $proud_3a$ | .004 |

Table 4: The highest ranked features for Extraversion on the ESSAYS dataset, averaged across the 10 cross-validation folds, using the $\chi^2$ feature selection.

**Improved impact of WSD**

In a more complex setting of WSD, we can partially resolve the issue mentioned above by (1) applying the *Selective.WSD* method and (2) combining WSD with semantic and/or sentiment information. Firstly, in Figure 4, we showed that the *Selective.WSD* method works better than the normal WSD method in selecting sense-level features for the APC. Especially, when we increase the number of topK features, the performance will drop. The reason for this difference was discussed in subsection 4.1. Secondly, we performed various experiments to show the benefit of combining WSD with semantic and sentiment features. Figure 2 indicates the differences between using WSD with semantic and/or sentiment features versus not-using WSD. Briefly, the combination of WSD with semantic and/or sentiment information works better in two cases of less-noise UGC data including ESSAYS and FACEBOOK on cEXT and cNEU personal trait. Our analysis shows that this is because cEXT and cNEU people use more pronoduns and emotional words than other personal traits.

## 6 Conclusion

This paper presents extensive experiments to explore the lexical-semantic resources on APC. Especially, WSD is combined with semantic and sentiment information to pose an improved performance in APC. In summary, we draw the following major conclusions. Firstly, using a dictionary (e.g., WordNet, WiktionaryEN) to remove noise-features often works well in most datasets. Secondly, applying WSD alone, in general, does not work in APC, especially on not-well-written UGC data. However, our proposed *Selective.WSD* works better than a basic WSD. Thirdly, applying WSD combining with semantic and/or sentiment features improve the performance for specific personal traits (i.e., cNEU, cEXT). Moreover, no personality specific resources are required in our method.

**Acknowledgments**

This work has been supported by the German Research Foundation under grant No. GU 798/14-1 and by Umeå University on federated database research.